\documentclass[11pt]{amsart}

\usepackage{lineno}
\usepackage{color}
\usepackage[]{graphicx}
\usepackage{amsmath}
\usepackage{amsfonts} 
\usepackage{amssymb}
\usepackage{theorem}
\usepackage{stmaryrd}

\begin{document}

\title{Atmospheric turbulence restoration by diffeomorphic image registration and blind deconvolution} 


\author{J\'er\^ome Gilles, Tristan Dagobert, Carlo De Franchis}
\address{DGA/CEP - EORD department, 16bis rue Prieur de la C\^ote d'Or, 94114 Arcueil Cedex, France}
\email{jerome.gilles@etca.fr,tristan.dagobert@etca.fr}

\maketitle

\begin{abstract}
A novel approach is presented in this paper to improve images which are altered by atmospheric turbulence. Two new algorithms are presented based on two combinations of a blind deconvolution block, an elastic registration block and a temporal filter block. The algorithms are tested on real images acquired in the desert in New Mexico by the NATO RTG40 group.
\end{abstract}

\section{Introduction}
\begin{figure}[ht!]
\begin{center}
\begin{tabular}{ccc}
\includegraphics[height=3.5cm]{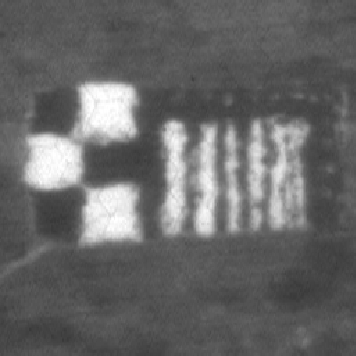} &
\includegraphics[height=3.5cm]{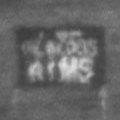} 
\end{tabular}
\end{center}
\caption{Example of images acquired by passive imagers during the NATO RTG40 campaign}
\label{fig:exturb}
\end{figure}

The image atmospheric turbulence distortion phenomenon remains a hard problem specially for horizontal ground imagers. Many works were done for astronomical images but the models used are not adapted for ground imaging systems. If these effects do not really affect an human observer in the case of weak turbulence, it can be very awkward for an automatic target recognition algorithm because the objects may be very different from the shapes learned by the algorithms. Some examples of different levels of turbulence are given in figure \ref{fig:exturb}. We can clearly see that the distortion affect the image comprehension (look at the letters on the right image). For a few years, a NATO Task Group (RTG40) focus its work on the validation of atmospheric turbulence models for infrared passive and active imaging systems. In order to evaluate the models, an acquisition campaign was done in the desert in New Mexico where different levels of turbulence were accessible. More generally the image distortion appears for different phenomena (for example underwater imaging systems are subject to the scattering effect which can be viewed as an image distortion). \\

In this paper, we adress some work in the field of image distortion restoration. The technique we propose modelize the alteration by a combination of image warping and blurring. Then we try to inverse the process by blind deconvolution and image registration.\\
First, in section \ref{sec:turb} we begin to describe the related phenomena and propose to use the model of Frakes et al. \cite{frakes2,frakes1}.
In section \ref{sec:filter}, we remind the temporal filters which will be used in the global restoration algorithm. In section \ref{sec:deconv}, we describe the blind deconvolution algorithm we use which comes from the work of Chan et al. (see \cite{chan1999,chan1998}). In section \ref{sec:diffeo}, we give details about the image registration which is based on the diffeomorphism approach described in the work of Youn\`es (\cite{younes2000,younes2001}) and Beg (\cite{beg}). Then, we present the architecture of the image distortion restoration algorithm in section \ref{sec:rest} and section \ref{sec:exp} show many experiments on images on which turbulence is present. Then we will conclude and give some perspectives of our work.

\section{Turbulence image distortion effect} \label{sec:turb}
If many works were done about the turbulence effect on astronomical images, the case of horizontal turbulence for ground to ground imagers is less studied. In \cite{lemaitre}, the authors try to use local filters (Wiener filtering, laplacian regularization,$\ldots$) but the local properties are obtained by a block partitionning of the image. The main drawback is that some block effects appear in the restored images. A first report is that no turbulence models exist for horizontal turbulence unlike for vertical turbulence. Many efforts are currently done around the world to get a good model which will be representative of the real effects. For example, the NATO Research Task Ground-40 (RTG40) deals with this kind of problem for active imaging systems. \\
If the modelization of the physical phenomenon is an important topic, the question of image (which is altered by turbulence) restoration is also an important question because it acts directly on the observation performances. The work presented in this paper is a part of the work we did in the RTG40 on image restoration.

As no specific models are available to use for image restoration, we decided to use the general model proposed by Frakes et al. \cite{frakes2,frakes1}. The authors propose to modelize the turbulence phenomenon by using two ``basic'' operators:

\begin{equation}\label{eq:turbulence}
I_{obs}=D(H(I_{ideal}))+b
\end{equation}

where $I_{obs}$ is the observed image, $H$ is a kernel of blur and $D$ is an operator which represents the geometric distortions operator caused by turbulence. In the rest of the paper, we will not take care about the noise $b$ because it can be considerably reduced by some simple temporal filtering (see section \ref{sec:filter}). In this paper, we explore a new way to retrieve a restored version $I_{rest}$, of an ideal image $I_{ideal}$ from the observed data $I_{obs}$ by using some operators which will approximate $H^{-1}$ and $D^{-1}$. We choose a blind deconvolution technique to approximate $H^{-1}$ and a registration algorithm for $D^{-1}$. Before examining these operators, we recall the temporal filters commonly used in the literature which will be used on the global restoration algorithm in the end of the paper.

\section{Temporal filters}\label{sec:filter}
In \cite{gilles2007,ove2006}, temporal filters are used to restore sequences acquired by the imaging systems. We write $\{I_n\}_{0 \leqslant n \leqslant N}$ a sequence, where $N$ is the number of frames in the sequence. We assume that images are size of $I\times J$ and the coordinate of one pixel is written $(i,j)$ (then the image domain is $\Omega=\llbracket 0;I \rrbracket\times \llbracket 0;J \rrbracket$). The temporal mean and temporal median filters are defined respectively by

\begin{equation}
\forall (i,j)\in\Omega \qquad I_{mean}(i,j)=\frac{1}{N}\sum_{n=0}^N I_n(i,j)
\end{equation}
and
\begin{equation}
I_{med}(i,j)= MED((I_n(i,j))_{0\leqslant n\leqslant N}) \quad \text{we assume that $N$ is odd}
\end{equation} 
We recall that this filter consists of rearranging the vector
\begin{equation}
\left(I_0(i,j), I_1(i,j), \ldots, I_{N-1}(i,j)  \right)
\end{equation}
in an increasing way, then we get the vector
\begin{equation}
\left(\tilde{I}_0(i,j), \tilde{I}_1(i,j), \ldots, \tilde{I}_{N-1}(i,j)  \right)
\end{equation}
where $\tilde{I}_n(i,j)\leqslant\tilde{I}_{n+1}(i,j)$. As $N$ is assumed to be odd, the median value is $\tilde{I}_{\frac{N-1}{2}}(i,j)$.\\

\begin{figure}[ht!]
\begin{center}
\includegraphics[scale=0.5]{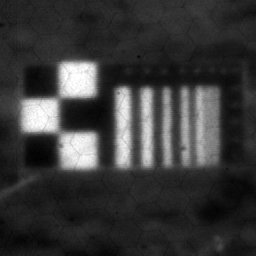}
\end{center}
\caption{Result obtained by temporal median filtering.}
\label{fig:median}
\end{figure}

In \cite{gilles2007}, the author prove by using a statistical argument that the median filter is a better choice than the mean filter and gives better results. An example of result obtained at the output of the median filter applied on a turbulenced sequence is given on figure \ref{fig:median}. The improvement that temporal filters gave are interesting but many experiments show that geometric distortions remain in the sequence. This kind of filters could be sufficient for an human observer but in the case of higher level of turbulence, an automatic target recognition (ATR) algorithm needs higher restoration performance rates.

\section{Blind deconvolution}\label{sec:deconv}
\begin{figure}[ht!]
\begin{center}
\begin{tabular}{|c|c|}
\hline
initial image &
blurred image \\
\includegraphics[width=0.25\textwidth]{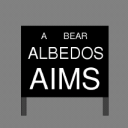}&
\includegraphics[width=0.25\textwidth]{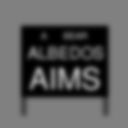}\\
\hline
deconvolved image &
kernel of blur\\
\includegraphics[width=0.25\textwidth]{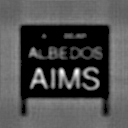}&
\includegraphics[width=0.25\textwidth]{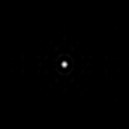}\\
\hline
\end{tabular}
\end{center}
\caption{Results of the blind deconvolution algorithm with $\alpha_1=10^{-5}$ and $\alpha_2=10^{-3}$.}
\label{fig:deconv}
\end{figure}

As we mentioned in section \ref{sec:turb}, we need to deconvolve images to decrease the blurring effect of the atmosphere.
We choose a blind deconvolution approach because in the case of atmospheric turbulence we never know the kernel of blur and the advantage of blind deconvolution techniques is that we don't need to know any a priori model about the kernel of blur. The algorithm estimates iteratively the deconvolved image and the kernel of blur. \\
In this section we describe the blind deconvolution algorithm we choose which is based on the work of Chan et al. \cite{chan1999,chan1998}.\\
If we denote $I_{blu}$ the blurred image and $H$ the kernel of blur then it is assumed that
\begin{equation}
I_{blu}=H\ast I_{ideal}+b
\end{equation}
where $b$ is an additive noise and $\ast$ is the convolution product. The algorithm of Chan et al., which estimates $I_{ideal}$ and $H$, is based on an iterative method. The authors assume that $I_{ideal}$ and $H$ are function taken in a space $V$ defined by
\begin{equation}
V=\left\{u\in L^2(\Omega)\; / \; \|u\|_2+|u|_{TV} \leqslant c\right\}
\end{equation}
where $c$ is a constant and the semi-norm $|u|_{TV}\triangleq \|\nabla u\|_1$. Then the authors propose to minimize the following functional to find $I_{ideal}$ and $H$:
\begin{equation}
E(I_{ideal},H)=\frac{1}{2}\|H\ast I_{ideal}-I_{blu}\|_2^2+\alpha_1\int_{\Omega}|\nabla I_{ideal}|dx+\alpha_2\int_{\Omega}|\nabla H|dx
\end{equation}
Then, if we apply the G\^ateaux derivative we obtain (see \cite{chan1999,chan1998} for details), $\forall x\in\Omega$
\begin{eqnarray}
 \frac{\partial E}{\partial H}(x)&=&I_{ideal}(-x)*(I_{ideal}*H-I_{blu})(x)
-\alpha_2\nabla\cdot\left( \frac{\nabla H(x)}{\vert\nabla
H\vert}\right)=0,\label{eq:AMuu}\\
 \frac{\partial E}{\partial I_{ideal}}(x)&=&H(-x)*(H*I_{ideal}-I_{blu})(x)
-\alpha_1\nabla\cdot\left( \frac{\nabla I_{ideal}(x)}{\vert\nabla
I_{ideal}\vert}\right)=0.
\end{eqnarray}
The authors suggest to use an iterative fixed-point method to solve this system of equations and they prove that an equivalent matrix formulation exists. These matrices are Toeplitz plus Hankel matrices and this property permits us to use an efficient diagonalization technique based on a discrete cosine transform.\\
Figure \ref{fig:deconv} shows the results we get when we apply this algorithm on the top-right image (this image is a simulated blurred version of the top-left image by a gaussian kernel). The bottom row gives the deconvolved image (on bottom-left) and the estimated kernel of blur (on bottom-right). We can see that the algorithm works well and gives a good estimation of the kernel.

\section{Image registration by diffeomorphism}\label{sec:diffeo}
\begin{figure}[ht!]
\begin{center}
\includegraphics[scale=0.5]{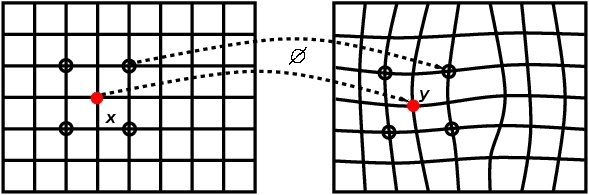}
\end{center}
\caption{Principle of image warping by diffeomorphism.}
\label{fig:warping}
\end{figure}

In this section, we adress the problem of the image distortions correction. We assume that image distortions are equivalent to geometric distortions and we choose to modelize it by a warping operator denoted $\phi$. Assume we want to warp an image $I_0$ to an image $I_1$, (see figure \ref{fig:warping}, where $I_0$ and $I_1$ are relatively close), a general energy formulation can be written
\begin{equation}
E(\phi)=D(\tilde{I}_0,I_1)+R(\phi)
\end{equation}

where $\tilde{I}_0=I_0\circ \phi$ is the warped version of $I_0$ to $I_1$, $D$ represents a similarity measure between $\tilde{I}_0$ and $I_1$ and $R$ is a regularization term to avoid the indesirable behaviour of $\phi$. Now the question is how to choose $D$ and $R$?\\

Many models exist in the literature, statistical models (\cite{hermosillo}), physical models (\cite{richard}), $\ldots$ In this paper we choose a warping technique based on diffeomorphisms which uses concepts commonly used in fluid mechanics and applied in medical image processing applications \cite{younes2000,younes2001,beg}.\\
First, we recall the definition of a diffeomorphism: given two manifolds $M$ and $N$, a bijective map $\phi$ from $M$ to $N$ is called a diffeomorphism if both $\phi$ and $\phi^{-1}$ are differentiable. In \cite{beg,younes2000,younes2001}, the authors prove that it exists a space $V$ of velocity fields defined by
\begin{equation}
V=\left\{v\in H^2(\Omega,\mathbb{R}^2)\cap L^2(\Omega,\mathbb{R}^2)\right\}
\end{equation}
embedded by the following inner product
\begin{equation}
\forall f,g \in V \qquad \langle f,g\rangle_V \triangleq \langle Lf,Lg\rangle_{L^2}
\end{equation}
where $L=-\alpha\Delta+\gamma$ is the Cauchy-Navier differential operator ($\alpha$ and $\gamma$ are parameters). Then for each time $t$ it exists a velocity field $v_t\in V$ such that 
\begin{equation}
\frac{d\phi_t}{dt}=v_t(\phi_t).
\end{equation}
The authors suggest to find this velocity field by minimizing the following energy term
\begin{eqnarray}
\hat{v}&=&\underset{v\in V}{\arg\min} E(v),\\
&=&\underset{v\in V}{\arg\min}\frac{1}{2}\int_0^T\Vert v_t\Vert_V^2dt + \frac{C}{2}\Vert I_0 \circ \phi_{T,0}^v -I_T\Vert_{L^2}^2.
\end{eqnarray} 
Then they calculate the corresponding Euler-Lagrange equation and use the Neumann's limits boundary condition to solve this problem (see \cite{beg} for the details). They also show that the inner product over $V$ can be easily calculated by filtering in the Fourier domain. All numerical aspects and the implementation details of this algorithm by using a gradient steepest descent method can be found in \cite{beg}.\\

\begin{figure}[ht!]
\begin{center}
\begin{tabular}{ccc}
\includegraphics[scale=0.3]{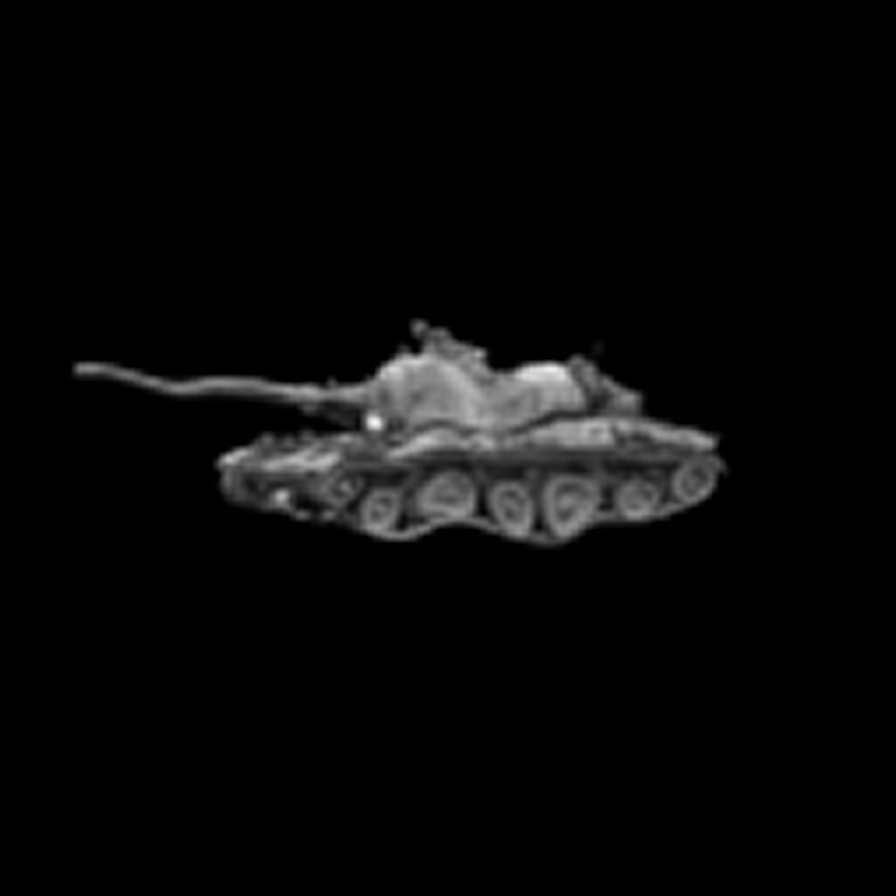} &
\includegraphics[scale=0.3]{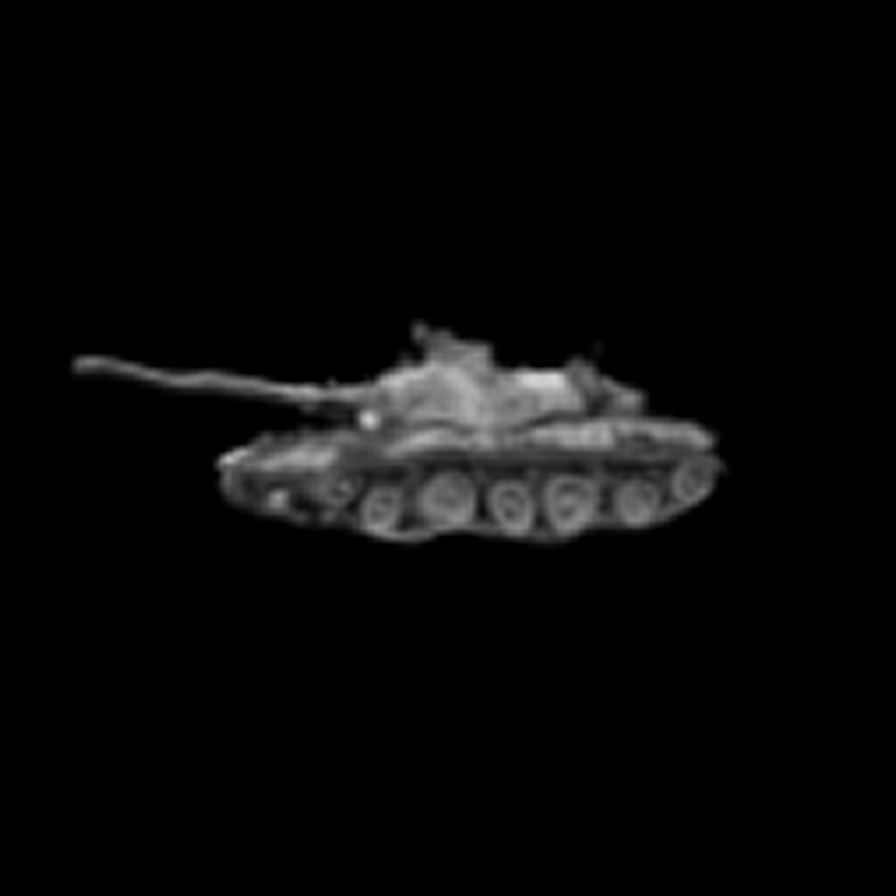} &
\includegraphics[scale=0.3]{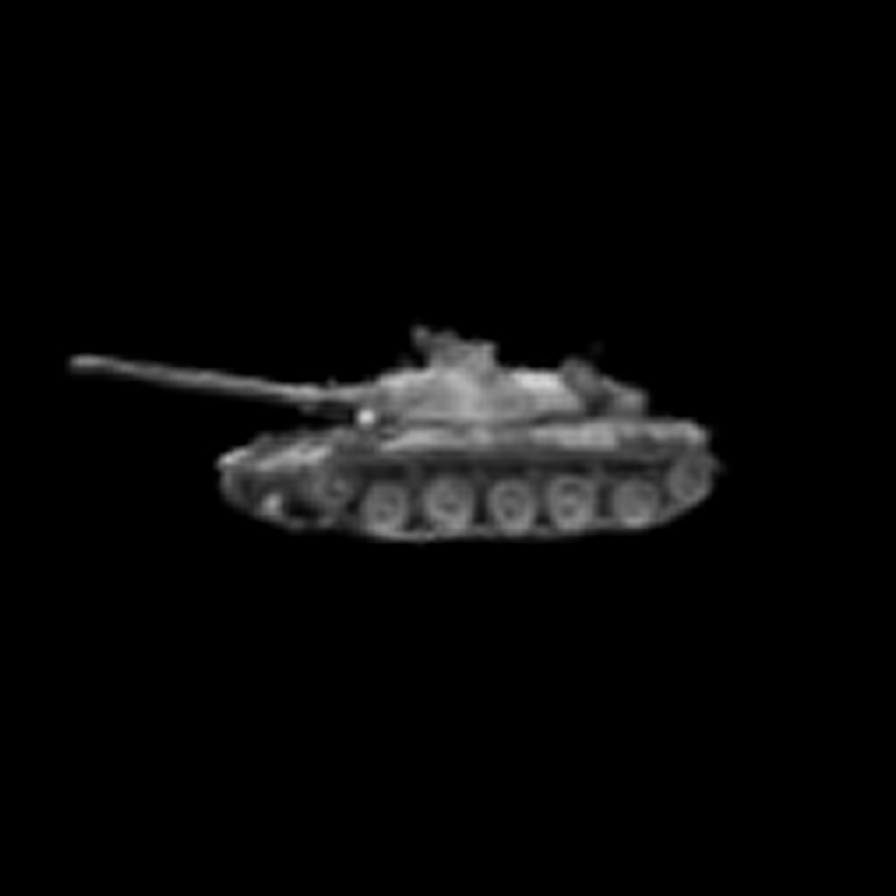} \\
Initial & Step 10 & Step 80
\end{tabular}
\end{center}
\caption{Results of warping algorithm by diffeomorphism.}
\label{fig:tank}
\end{figure}

In figure \ref{fig:tank}, we can see the evolution during the iteration time of the algorithm on simulated distortions (we use the original image as reference for the diffeomorphism). We can see that the algorithm works very well, the geometric distortions are removed.\\
We also study the influence of the parameters (see \cite{dago}) $\alpha$ and $\gamma$ on different basic distortions and we can conclude that we must choose these parameters under the following constraints: $(\alpha,\gamma)\in [0.01,0.3]\times [0.1, 1]$ with $\alpha<\gamma$.

\section{Image distortion restoration}\label{sec:rest}
In this section, we adress the global image alteration restoration problem as defined in the turbulence case in section \ref{sec:turb}. In this section, we assume that we have a sequence of images of the observed scene $\{I_n\}_{0 \leqslant n \leqslant N}$ where $N$ is the number of frames in the sequence. As we mentioned previously, we will use the blind deconvolution algorithm to evaluate $H^{-1}$ and the diffeomorphism registration technique to evaluate $D^{-1}$. Two iterative methods can be retained to get the restored images. The first one, we call it FRD algorithm, is based on the theoretical model described in section \ref{sec:turb} where $I_{rest}=D^{-1}(H^{-1}(I_{obs}))$ ($I_{rest}$ is the restored image). The second, we call DFR algorithm, is based on the practical approach proposed by Frakes et al. \cite{frakes1,frakes2} which corresponds to $I_{rest}=H^{-1}(D^{-1}(I_{obs}))$. Before to detail these algorithms, an important question arises: which reference can we use for the diffeomorphism?  \\

\begin{figure}[ht!]
\begin{center}
\includegraphics[width=\textwidth]{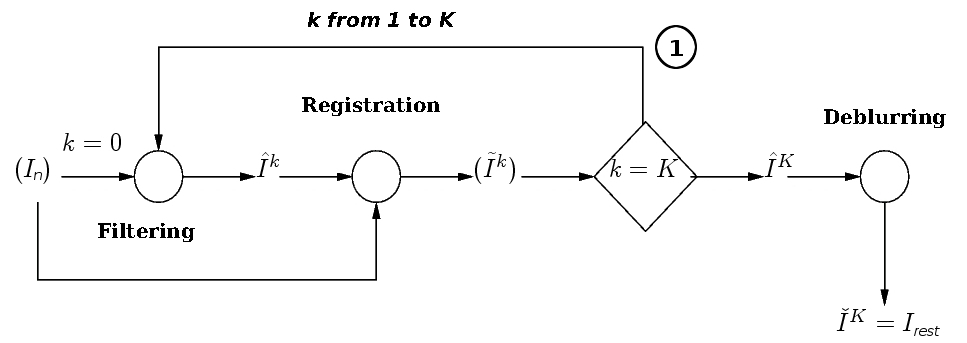}
\end{center}
\caption{Global restoration algorithm based on the theoretic model (FRD algorithm).}
\label{fig:frd}
\end{figure}

\begin{figure}[ht!]
\begin{center}
\includegraphics[width=\textwidth]{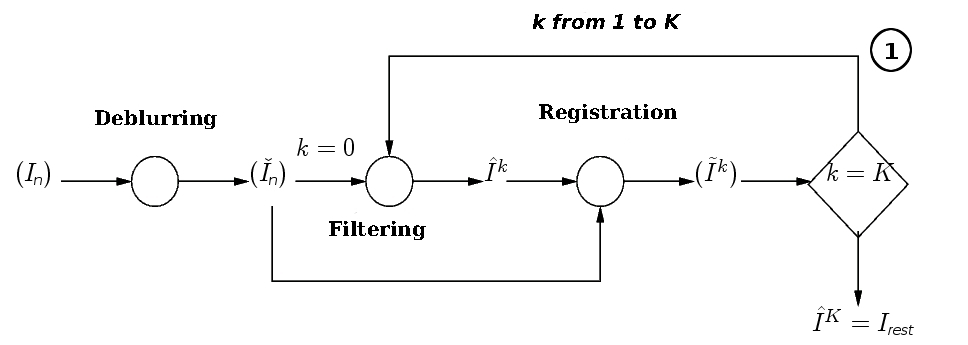}
\end{center}
\caption{Global restoration algorithm based on the practical approach of Frakes (DFR algorithm).}
\label{fig:dfr}
\end{figure}

As we see in section \ref{sec:filter} the temporal filters could be some good candidates. So we decided to use these temporal filters to get a reference image as input of the diffeomorphism registration technique.\\
The principle of the FRD algorithm is shown in figure \ref{fig:frd}. The input sequence is used by the temporal filter to generate a first reference $\hat{I}^k$ for $k=0$. The input sequence is then warped on this reference. This step gives us a registrated sequence $(\hat{I}^k)$. Then we can get a new reference image from $(\hat{I}^k)$ and iterate the process (\textcircled{1}). After $K$ iterations, we apply the blind deconvolution algorithm on the last reference image to get the final result $I_{rest}$.\\
The DFR algorithm is summarized in figure \ref{fig:dfr}. The main difference with the FRD algorithm is that the blind deconvolution algorithm is first applied on each frame of the input sequence. And the final result is simply the last reference image $\hat{I}^K$. 

\section{Experiments}\label{sec:exp}
\begin{figure}[ht!]
\begin{center}
\begin{tabular}{ccc}
\includegraphics[scale=0.43]{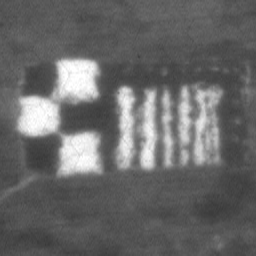} &
\includegraphics[scale=0.43]{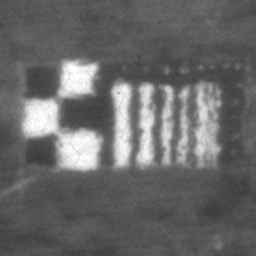} &
\includegraphics[scale=0.43]{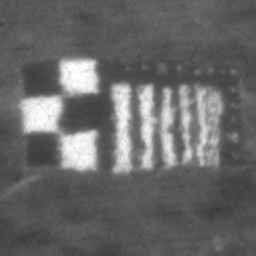} \\
\end{tabular}
\end{center}
\caption{Example of frames extracted from the real test sequences.}
\label{fig:realtest_init}
\end{figure}

\begin{figure}[ht!]
\begin{center}
\begin{tabular}{ccc}
\includegraphics[scale=0.43]{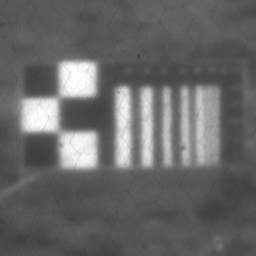} &
\includegraphics[scale=0.43]{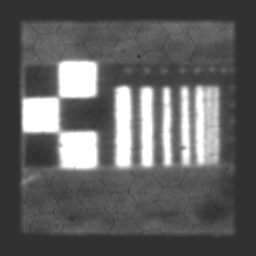} &
\includegraphics[scale=0.43]{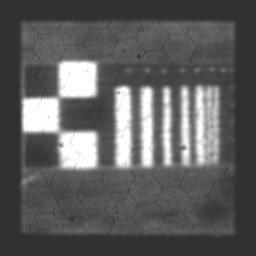} \\
(a) & (b) & (c)
\end{tabular}
\end{center}
\caption{Results on passive imagery: (a) temporal median image output, (b) FRD result, (c) DFR result.}
\label{fig:realtest_pa}
\end{figure}

In this section, we present some results we get by testing the FRD and DFR algorithms. We use a sequence acquired by a passive imager during the NATO RTG40's campaign in the desert in New Mexico. Some frames of the test sequence are shown in fig.\ref{fig:realtest_init}. Then, we apply the FRD and DFR algorithms on these sequences (we use only one iteration, $K=1$, $\alpha_1=2.10^{-2}, \alpha_2=1, \alpha=0.01, \gamma=0.7$). The results and the temporal median filter output are shown on fig.\ref{fig:realtest_pa}. We can clearly see that the proposed algorithms give the best results, the distortion are removed by the registration and the blind deconvolution permits to retrieve some high frequencies of the images (the edges are sharpened, the white dots are more observable and the hexagonal pattern due to the fiber network of the sensor is also restored). The DFR algorithm seems to have a better accuracy than the FRD algorithm (the contrasts between the objects are better enhanced).

\section{Conclusion - Futur work}\label{sec:concl}
In this paper, we present a novel approach to restore the atmospheric turbulence effect on video sequence. Two combinations of an elastic image registration algorithm and blind deconvolution are used in concordance with the physical model proposed by Frakes et al. \cite{frakes1,frakes2}. We perform our algorithms to real data acquired during the NATO RTG40 campaign. Our algorithm outperfom the results we get by commonly used temporal filters, the object's geometry is regularized and some high frequencies are retrieved. These characteristics improve the visual identification performance rates and will be necessary for an automatic target recognition algorithm.\\
We plan to explore the capabilities of the two algorithms in future work. As mentioned in section \ref{sec:rest}, the process can be iterated then we plan to test the influence of all parameters ($\alpha_1, \alpha_2, \alpha, \gamma, K$) on many sequences (in passive and active imaging). This work will be done in a future paper. Another theoretically interesting way of research could be unify the deconvolution process inside the registration algorithm and then create only one algorithm which could be more fast.

\bibliographystyle{plain}
\bibliography{diffeo}

\end{document}